# Intuitionistic Fuzzy Cognitive Maps for Interpretable Image Classification

Georgia Sovatzidi[1], Michael D. Vasilakakis[1], and Dimitris K. Iakovidis[1,2]

[1] Department of Computer Science and Biomedical Informatics, University of Thessaly, Papasiopoulou St. 2-4, 35131 Lamia, Greece

[2] Corresponding Author (diakovidis@uth.gr)

## Abstract

Several deep learning (DL) approaches have been proposed to deal with image classification tasks. However, despite their effectiveness, they lack interpretability, as they are unable to explain or justify their results. To address the challenge of interpretable image classification, this paper introduces a novel framework, named Interpretable Intuitionistic Fuzzy Cognitive Maps ($I^2$FCM). Intuitionistic FCMs (iFCMs) have been proposed as an extension of FCMs offering a natural mechanism to assess the quality of their output through the estimation of hesitancy, a concept resembling human hesitation in decision making. In the context of image classification, hesitancy is considered as a degree of unconfidence with which an image is categorized to a class. To the best of our knowledge this is the first time iFCMs are applied for image classification. Further novel contributions of the introduced framework include the following: a) a feature extraction process focusing on the most informative image regions; b) a learning algorithm for automatic data-driven determination of the intuitionistic fuzzy interconnections of the iFCM, thereby reducing human intervention in the definition of the graph structure; c) an inherently interpretable classification approach based on image contents, providing understandable explanations of its predictions, using linguistic terms. Furthermore, the proposed $I^2$FCM framework can be applied to DL models, including Convolutional Neural Network (CNN), rendering them interpretable. The effectiveness of $I^2$FCM is evaluated on publicly available datasets, and the results confirm that it can provide enhanced classification performance, while providing interpretable inferences.

1. ## Introduction

Fuzzy Cognitive Maps (FCMs) are graph-based knowledge models composed of concepts and causal relationships among them (Kosko, 1986). Due to their simple yet efficient structure, FCMs have been increasingly used for many applications of various scientific fields, including engineering, medicine, and business fields (Orang et al., 2023). However, the potential of FCMs has not been yet sufficiently investigated with respect to image-based decision making. A relevant study (Subramanian et al., 2015) proposed an FCM with a Hebbian learning algorithm developed to predict breast cancer risk using mammographic image features. In (Hilal et al., 2022), FCMs were used to classify remote sensing image scenes, while a swarm intelligent algorithm was used for parameter tuning. Preliminary works of the authors exploited FCMs to develop interpretable image classifiers. More specifically, the xFCM, proposed in (Sovatzidi et al., 2022a), was a first approach of an auto-constructed FCM capable of classifying images. Furthermore, the Interpretable FCM-based Feature Fusion (IF³) framework proposed in (Sovatzidi et al., 2022b) enabled the fusion of different features extracted using various methods and adapted for automated lesion risk assessment in medical images. However, the interpretations derived from these approaches were considering each image as a whole, neglecting local information from image regions that encompasses details about their content.

Over the years, various modifications have been proposed to overcome the limitations of FCMs (Felix et al., 2019; Schuerkamp and Giabbanelli, 2023). Such limitations include the participation of experts for the definition of the graph structure that may introduce biases, thus reducing the accuracy of the models. Considering the hesitation that experts may have when defining the relations among the graph concepts, iFCM-I and iFCM-II were introduced. These models can effectively tackle uncertainty by incorporating intuitionistic fuzzy sets (IFSs) (Iakovidis and Papageorgiou, 2010; Papageorgiou and Iakovidis, 2012). They have demonstrated a more robust decision-making performance compared to the conventional FCM models, while offering a natural mechanism to assess the quality of their output through hesitancy estimation. Despite their advantages, to date, only a limited number of applications have been based on iFCMs. These include pneumonia severity assessment (Iakovidis and Papageorgiou, 2010), chemical process control (Papageorgiou and Iakovidis, 2012), time series forecasting (Hajek et al., 2020; Orang et al., 2023), supplier selection tasks (Hajek and Froelich, 2019), and celiac disease prediction (Amirkhani et al., 2018).

In this paper, we introduce a novel framework based on iFCMs for image classification, which is motivated by the need for robust, uncertainty-tolerant interpretable ML models. To the best of our knowledge, it is the first time that iFCMs are used in this context. The contributions of this paper include:

- A CNN-based feature extraction process focusing on the most informative image regions.
- A methodology for data-driven construction of IFSs.
- A learning algorithm enabling data-driven determination of the interconnections of the graph are determined automatically, based on the defined IFSs.
- An inherently interpretable classification approach exploiting the iFCM structure for the explanation of the classification outcomes based on local information derived from image regions.

The remainder of this paper is organized as follows: Section 2 provides a literature review of the state-of-the-art interpretable image classification models. Section 3 recalls prior knowledge on conventional and intuitionistic FCMs. The proposed framework is presented in Section 4. Experiments conducted in this study are included in Section 5. Finally, conclusions as well as future research directions are discussed in Section 6.

2. Related Work

Machine learning models are increasingly being adopted to make decisive and meaningful predictions in several critical tasks of various scientific fields (Angelov et al., 2021). However, the absence of interpreting the results in an intelligible way results in machine learning (ML) approaches being considered in some cases unreliable and "black boxes" (Ding et al., 2022). In recent years, there have been increasing efforts to develop image classification models explaining their inferences (Ibrahim and Shafiq, 2023). Among them, methods based on fuzzy logic offer the advantage of being robust under uncertainty and imprecision, while enabling human-like non-numerical knowledge representation and reasoning.

Recently, the explainable Deep Neural Network (xDNN) was proposed as a solution to explain internal architecture of deep learning models (Angelov and Soares, 2020a). The xDNN model is based on the use of prototype learning, where prototypes are actual training images derived from the local peaks of the feature space density. This density is defined by using pre-trained deep neural network, allowing automatic extraction of more abstract and discriminative high-level features from the training images. Deep Machine Reasoning (DMR) extended xDNN to address more complex multi-class problems (Angelov and Soares, 2020b). DMR integrates prototype learning with decision tree-based inference while addressing class imbalances by generating synthetic data around the prototypes based on the training data. The decision-making process of the DMR model is based on pairwise class comparisons providing insights into the image classification. The interpretable Deep Rule-Based (DRB) classifier was proposed for image recognition tasks (Angelov and Gu, 2018). Specifically, DRB identifies image prototypes from the empirically observed image data, extracts features from image prototypes using pre-trained deep neural network and groups these features into clusters for each class. It then constructs a data-driven set of *IF…THEN…* fuzzy rules that can be updated continuously without a full retraining. The rule-base is formed of zero-order autonomous simplified fuzzy rules. However, it is frequently observed that the inner structure of DRB can become overly complex, reducing the interpretability for humans particularly in large-scale image classification problems. To tackle this issue, Hierarchical Deep Rule-Based Classifier (H-DRB) incorporates a tree structural model, where the large number of prototypes in the training data are aggregated into smaller groups of more descriptive prototypes(Gu and Angelov, 2020). The structure of H-DRB rule-based model focuses on breaking down the decision-making process into a structured set of rules, providing a clearer understanding of model decisions compared to DRB.

Table 1 summarizes the features of the current interpretable fuzzy image classification models in comparison to I²FCM. The latter, like the other interpretable models, derives interpretations by exploiting similarities between images. However, although all of them consider the similarities to explain class memberships, only the FCM-based methods use fuzzy sets in the reasoning process. In this way, they offer insights about the image classification results that can be easier perceived through linguistic expressions. Furthermore, a distinct advantage of I²FCM compared to the rest of the methods is that it utilizes local information from image regions, including semantically relevant details about image contents; thus, offering enhanced explanations.

**Table 1** Interpretable Fuzzy Image Classification Models

| Model | Image Similarities | Linguistic Interpretation | Local Information | Method |
|---|---|---|---|---|
| xDNN | ✓ | – | – | Fuzzy Rules |
| H-DRB | ✓ | – | – | |
| DRB | ✓ | – | – | |
| DMR | ✓ | – | – | |
| xFCM | ✓ | ✓ | – | FCM |
| **I²FCM** | ✓ | ✓ | ✓ | |

## 3. Preliminaries
### 3.1 Fuzzy Cognitive Maps

FCMs constitute a powerful tool for complex system modeling (Kosko, 1986). They are graph-based models, composed of nodes and weighted arcs. The nodes represent concepts related to the complex system being modeled, and the arcs represent rules expressing causal relationships between these concepts. Each of the *N* concepts $C_i, i = 1, ..., N$, of the FCM has a value $A_i \in [0,1]$. The weight of an arc between two concepts $C_j, j = 1, ..., N$, and $C_i, i \neq j$ denoted as $w_{ji} \in [-1,1]$, represents the degree to which $C_j$ influences $C_i$. There are three types of causal relationships: a) positive ($w_{ji} > 0$), which means that an increase in the value of $C_j$, causes an increase of the value of $C_i$, b) negative ($w_{ji} < 0$), indicating that an increase in the value of $C_j$, causes a decrease of the value of $C_i$, and c) neutral ($w_{ji} = 0$), meaning that there is no relationship between $C_j$ and $C_i$. To construct an FCM, a number of concepts along with the respective weighted arcs connecting them, are usually defined manually with the contribution of domain experts. Once the graph is constructed, given a test case, the reasoning phase takes place until the FCM converges to a steady state. The values of the output concepts retrieved from that state represent its inferences. During the reasoning process the concept values $A_i \in [0,1]$ are iteratively calculated as follows (Kosko, 1986):

$$A_i^{t+1} = f(A_i^t + \sum_{j=1, j \neq i}^{N} A_j^t \cdot w_{ji}) \quad (1)$$

where $A_i^{t+1}$ represents the value of $C_i$ at the iteration $t+1$, $w_{ji}$ is the influence of $C_j$ on $C_i$, and $f$ is a sigmoid function such as the log sigmoid, which maps the concept values within $[0,1]$(Bueno and Salmeron, 2009).

$$f(x) = \frac{1}{1+\exp(-x)} \quad (2)$$

The initial state vector $A^0$ represents the initial concept values, for $t = 0$.

### 3.2 Intuitionistic Fuzzy Cognitive Maps

Intuitionistic fuzzy sets (IFSs) (Atanassov, 1999) extend fuzzy sets by introducing the concept of non-membership as not necessarily a complement to the membership. Given a universe of discourse $G$, an IFS is defined as:

$$S = \{\langle x, \mu_S(x), \gamma_S(x)\rangle | \, x \in G\} \quad (3)$$

where $\mu_S(x) \in [0,1]$ and $\gamma_S(x) \in [0,1]$ define the degree of membership and non-membership, respectively of $x \in G$ to $S \subset G$. The hesitancy $h_S(x)$ of an element $x \in G$ to $S \subset G$ is defined as follows:

$$h_S(x) = 1 - \mu_S(x) - \gamma_S(x) \quad (4)$$

characterizing the indeterminacy (uncertainty) of the membership of $x$ in $S$.

In (Papageorgiou and Iakovidis, 2012), iFCM-II was proposed as an extension of the original FCM model exploiting IFSs to model the uncertainty in the determination of the concept values and the weights of the model, respectively. An example of an iFCM-II is illustrated in Fig. 1, where IFSs are denoted with the simplified notation $\{\langle v^\mu, v^\gamma\rangle\}_i, i = 1, \ldots, N$ and $\{\langle w^\mu, w^\gamma\rangle\}_{ji}, j = 1, \ldots, N$.

Uncertainty is modeled by the hesitancy of the IFSs, which is considered to represent the hesitation of humans in decision making. Given a constructed iFCM-II with determined weight pairs $\langle w^\mu, w^\gamma\rangle$, the reasoning process is performed by iteratively calculating the pairs $\langle v^\mu, v^\gamma\rangle$ of each $C_i, i = 1, \ldots, N$ using the following equation which iteratively updates the respective IFSs:

$$\{\langle v^\mu, v^\gamma\rangle\}_i^{t+1} = F\left(\{\langle v^\mu, v^\gamma\rangle\}_i^t \oplus \left(\bigoplus_{\substack{j=1 \\ j \neq i}}^{N} \left(\{\langle v^\mu, v^\gamma\rangle\}_j^t \otimes \{\langle w^\mu, w^\gamma\rangle\}_{ji}^t\right)\right)\right) \quad (5)$$

where the symbols "$\oplus$" and "$\otimes$", correspond to the summation and multiplication operators, respectively and $F(S) = \{\langle f_\mu, f_\gamma\rangle\}$ are transformation functions defined on IFSs.

As proved in (Papageorgiou and Iakovidis, 2012), a pair $\langle v^\mu, v^\gamma\rangle$ from $\{\langle v^\mu, v^\gamma\rangle\}_i^{t+1}$ is calculated as follows:

$$(v_i^\mu)^{t+1} = f_\mu\big((v_i^\mu + (1 - v_i^\mu) \cdot \sigma_{iN})^t\big) \tag{6}$$

$$(v_i^\gamma)^{t+1} = f_\gamma\left(\left(v_i^\gamma \cdot \prod_{\substack{j=1 \\ j \neq i}}^{N}\big(v_j^\gamma + w_{ji}^\gamma - v_j^\gamma w_{ji}^\gamma\big)\right)^t\right) \tag{7}$$

where $(v_i^\mu)^{t+1}$ and $(v_i^\gamma)^{t+1}$ represent the membership and non-membership of $C_i$, respectively, at iteration $t+1$, and the calculation of $\sigma_{iN}$ is performed by

$$\sigma_{ij} = \begin{cases} v_1^\mu \cdot w_{1i}^\mu, & j = 1 \\ \sigma_{i(j-1)} + v_j^\mu \cdot w_{ji}^\mu - \sigma_{i(j-1)} \cdot v_j^\mu \cdot w_{ji}^\mu, & j > 1 \end{cases} \tag{8}$$

The variables $w_{ji}^\mu$ and $w_{ji}^\gamma$ represent the membership and non-membership, respectively, of the weights corresponding to the arcs directed from node $j$ to node $i$. The real hesitancy ($\hbar_i$) of a concept $C_i$ is given by (9):

$$\hbar_i = 1 - f_\mu^{-1}(v_i^\mu) - f_\gamma^{-1}(v_i^\gamma) \tag{9}$$

where $F^{-1}(S) = \{\langle f_\mu^{-1}, f_\gamma^{-1} \rangle\}$ represents the inverse of function $F(S)$, and $f_\mu^{-1}(v_i^\mu)$ and $f_\gamma^{-1}(v_i^\gamma)$, considering $f^{-1}$ to be the inverse of $f$, represent the real membership and non-membership values of that concept.

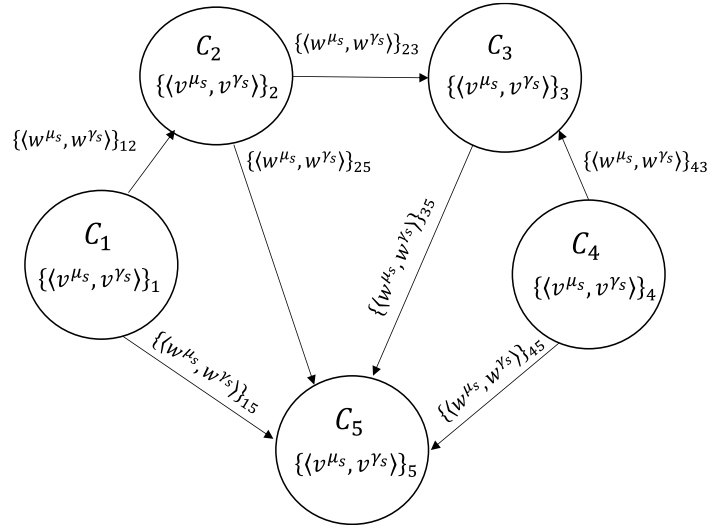

**Fig. 1.** An example of a five-concept iFCM model.

### 4. Methodology

The proposed framework for interpretable image classification involves two phases, a training, and a testing phase, which are detailed in the following paragraphs.

#### 4.1 Training Phase

*Step 1: Feature Extraction*

Let us consider a set of *K* training input images $I_k, k = 1, \ldots, K$, with ground-truth class vectors $\mathbf{y}_k = [y^1, y^2, \ldots y^F]$, where $F$ represents the total number of classes in the training dataset. An image belongs to class *f*, if $y^f = 1$, otherwise it does not belong to that class.

Firstly, the SLIC superpixel segmentation algorithm (Achanta et al., 2012) is used to segment $I_k, k = 1, \ldots, K$, into $P$ superpixels $p = 1, \ldots, P$, in line with (Liu et al., 2015). The generated superpixels constitute a superpixel map of height $\zeta$, and width $\rho$ (Fig. 2). The training input image $I_k, k = 1, \ldots, K$, is fed into a CNN to produce $\delta$ feature maps $\Delta_k \in \mathbb{R}^{\zeta \times \rho}$. The extracted feature maps are rescaled to the resolution of the superpixel map of $I_k, k = 1, \ldots, K$ (Fig. 2). A feature vector of the form $d_k^p$ is extracted from each superpixel $p = 1, \ldots, P$ of $I_k$, using average pooling. In the example of Fig. 2, a total of $P = 16$ feature vectors, $d_k^1, \ldots, d_k^{16}$, are obtained by following this process.

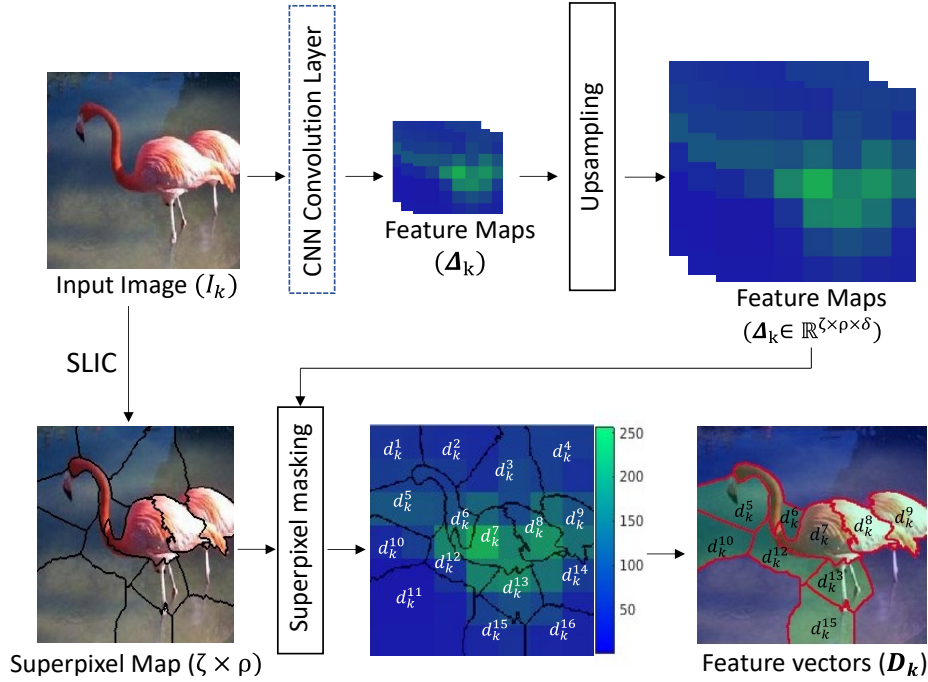

**Fig. 2.** The feature extraction process (step 1).

Within the I²FCM framework, a further selection of the most informative superpixels is introduced, considering their spatial interrelations in the superpixel map. Ideally, the most informative superpixels are those that describe the object represented in an image, and are more likely to be close to each other, while the superpixels that describe the background are arbitrarily placed at different positions (Hartley et al., 2021). To determine the superpixels that are likely informative, the following distances are calculated between all superpixels: a) their spatial distance based on their center of mass, and b) the distance among the feature vectors $d_k^p$ extracted from them. The superpixels having both of these distances smaller than the respective average distances are selected. In the example illustrated in Fig. 2, a subset of $P' = 9$ out of the total $P = 14$ superpixels, are selected (denoted a with a red outline). The set of feature vectors extracted from these superpixels is denoted as $D_k = \{d_k^1, \ldots, d_k^{P'}\}$.

*Step 2: Concept Mining*

The feature vectors $D_k = \{d_k^1, \ldots, d_k^{P'}\}, k = 1 \ldots K$ extracted from each training image $I_k$ belonging to class $f$, are grouped into $M_f$ clusters, using a conventional clustering algorithm, such as the *k*-means. The clusters facilitate a higher-level representation of the image content incorporating conceptual information. The medoids of the clusters are used as a more compact representation of this information (dashed squares in Fig.3–step 2). The resulting medoids are denoted as $r_m, m = 1, \ldots, \mathcal{M}$, where $\mathcal{M} = \sum_{f=1}^{F} M_f$. For each medoid $r_m$, a respective class vector $y_m = [y^1, y^2, \ldots y^F]$ is used to characterize its belongingness to a class, where $y^f \in \{0,1\}$, and $F$ is the total number of classes. Considering that the mined clusters with medoids $r_m$ represent concepts that exist in the input image, they will be considered as input concept nodes of the iFCM, which is constructed at a later step of this methodology (step 5). In Fig. 3 (step 2), images from three different classes ($F = 3$) are provided as an example, together with the features extracted from them during step 1. In this example, the medoids $r_1$ and $r_2$ correspond to class 1, $r_3$ and $r_4$ correspond to class 2, and $r_5$ and $r_6$ correspond to class 3.

*Step 3: Similarity Calculation*

In the third step of the training phase the similarity among the calculated medoids $r_m$, $m = 1, \ldots, \mathcal{M}$ and the extracted set of feature vectors $D_k = \{d_k^1, \ldots, d_k^{P'}\}, k = 1 \ldots K$ from a training image $I_k$ is calculated as follows (Vasilakakis and Iakovidis, 2023)

$$z_k(r_m) = \sum_{p=1}^{P'} \left(1 - \frac{\|d_k^p - r_m\|}{\sum_{i=1}^{\mathcal{M}} \|d_k^p - r_i\|}\right) \quad (10)$$

where $\|\cdot\|$ is a distance metric, *e.g.*, the Euclidean distance. The rationale of (10) is that it provides a normalized estimation of similarity, in the same way that a human recognizes and classifies objects based on comparisons with previous known related prototypes (Vasilakakis and Iakovidis, 2023). This process is illustrated with an example in Fig. 3 (step 3). The calculated similarities are then summed up and normalized within the interval [0, 1]. The resulting similarity vectors are depicted in the form of a histogram with their components appearing as color bars. Under each bin of these histograms a visual example of the respective medoid is provided. The different colors represent the three different classes included in the dataset.

*Step 4: Intuitionistic Fuzzy Modelling*

The fourth step includes the construction of IFSs linguistically characterizing the similarity of an input image with the images belonging to the target and other classes. The IFSs are modeled using a membership and a non-membership function. The membership function expresses the similarity of the feature vectors extracted from the input image with the medoids of the clusters derived from the training images of the class it belongs to. The non-membership function expresses the similarity of the feature vectors extracted from the input image with the medoids of the clusters derived from the images belonging to the other classes. More specifically, the membership functions $w^\mu(z_k(r_m))$ of the IFSs, are created by considering the similarities between $r_m$ and $D_k$, calculated for the images $I_k$ belonging to the same class ($y_k = y_m$). The non-membership functions $w^\gamma(z_k(r_m))$ of the IFSs, are created by considering the similarities between $r_m$ and $D_k$ calculated from images $I_k$ belonging to different classes ($y_k \neq y_m$). These IFSs are created by combining different fuzzy sets $\tilde{B}_m^e, e = 1, \ldots, E_b$, and $\tilde{Q}_m^e, e = 1, \ldots, E_q$ representing different degrees of membership and non-membership, respectively, as follows: initially the calculated similarities for the construction of $\tilde{B}_m^e$ are clustered into $E_b$ clusters, and the calculated similarities for the construction of $\tilde{Q}_m^e$ are clustered into $E_q$ clusters, using a conventional clustering algorithm, such as the *k*-means. The respective medoids $b_m^e, e = 1, \ldots, E_b$, and $q_m^e, e = 1, \ldots, E_q$ obtained from this clustering process are used to construct fuzzy sets with membership functions defined in the universe of similarities [0, 1], topping at these medoids. An example is illustrated in Fig. 3 (step 4). It shows the construction of fuzzy sets $\tilde{B}_m^e$ using the similarities $z_k(r_1), z_k(r_2)$ of $D_k = \{d_k^1, \ldots, d_k^{P'}\}, k = 1 \ldots K$ with $r_1$ and $r_2$ respectively, from the images belonging to the same class ($y_k(y^1) = y_m(y^1), m = 1,2$). Using the calculated medoids $b_1^1, b_1^2, b_1^3$ three fuzzy sets $\tilde{B}_1^1, \tilde{B}_1^2, \tilde{B}_1^3$ are defined. In this example triangular membership functions are used for simplicity; however, membership functions of other shapes, *e.g.*, a Gaussian, are applicable. The membership functions are designed so that they top at these medoids (framed within a blue box in Fig. 3–step 4).

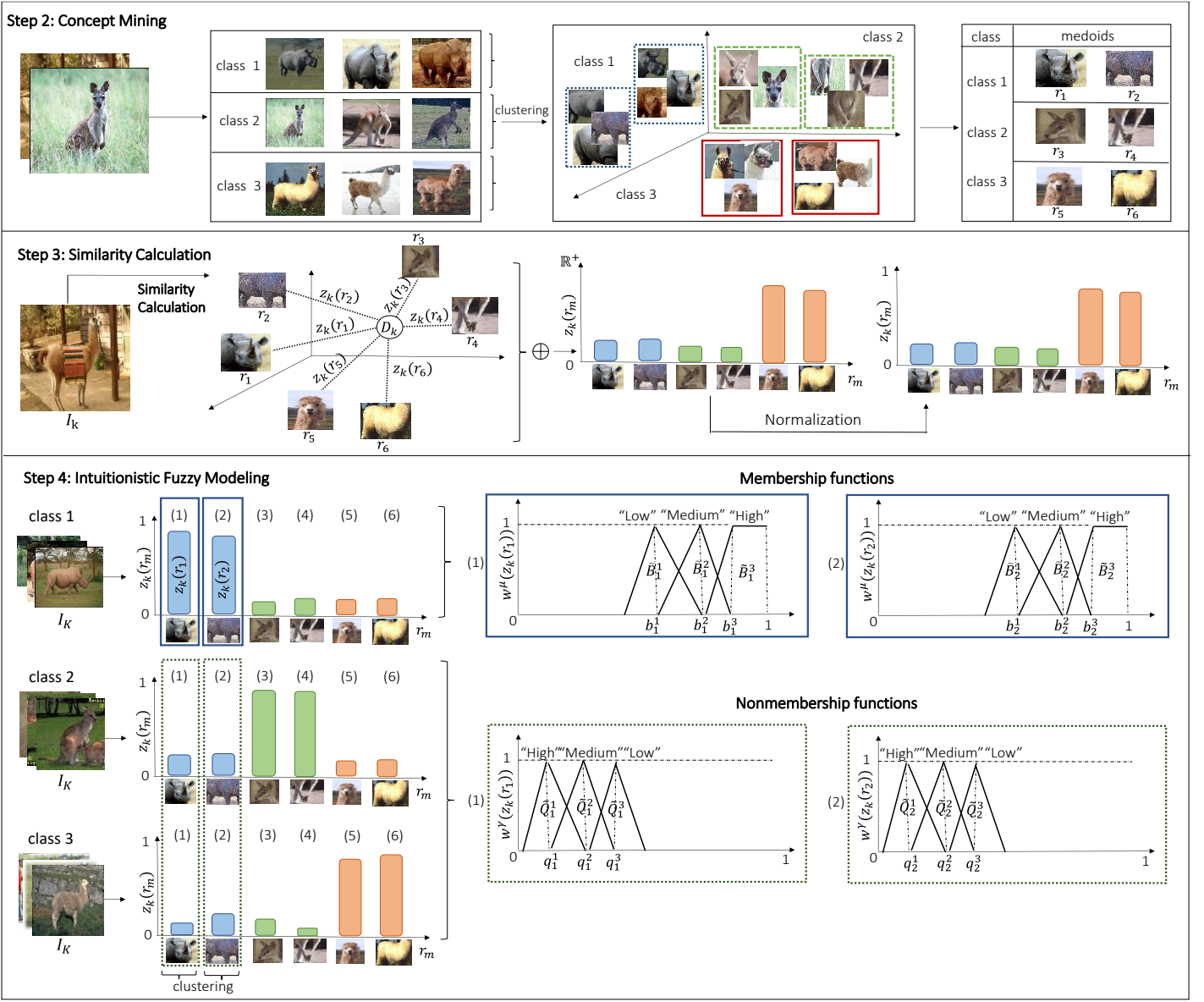

**Fig. 3.** Concept mining (step 2), similarity calculation (step 3), and intuitionistic fuzzy modelling (step 4).

For the construction of $\tilde{Q}_m^e$, the similarities $z_k(r_1)$, $z_k(r_2)$ of $D_k = \{d_k^1, \ldots, d_k^{P'}\}, k = 1 \ldots K$ with $r_1$ and $r_2$ respectively, are calculated from images belonging to different classes ($y_k(y^f) \neq y_m(y^1)$, $f = 2,3, m = 1,2$). The respective medoids $q_1^1, q_1^2, q_1^3$ are then used to construct $\tilde{Q}_1^1, \tilde{Q}_1^2, \tilde{Q}_1^3$, also considering membership functions of triangular shape (framed within a green box in Fig. 3–step 4). When using triangular functions, the intermediate triangles are extended to the nearest calculated centroid. Considering the non-membership functions, the intervals of the leftmost triangle extend to zero, corresponding to the "*High*" linguistic value, while the interval of the rightmost membership function is extended to one. Specifically, the rightmost triangle is transformed into a trapezoidal function with its right side extending to the nearest centroid (Vasilakakis and Iakovidis, 2023) (Fig. 3-step 4). In addition, the fuzzy sets have to be defined in such a way to overlap covering the whole range $[0,1]$, with no gaps. The overlap between the fuzzy sets aims to maintain continuity in the output.

In line with (Iakovidis and Papageorgiou, 2010), the subset of pairs of fuzzy sets from $\tilde{B}_m^e \times \tilde{Q}_m^e$ that satisfy

the conditions imposed by the definition of an IFS (section 2.2) are considered as pairs of membership

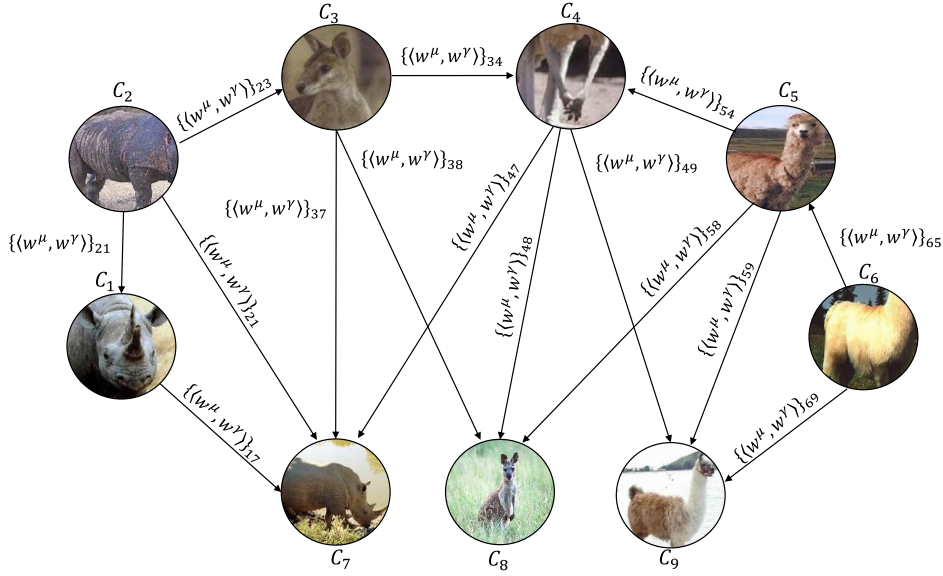

**Fig. 4.** IFCM example model for image classification.

and non-membership functions, generating a set of IFSs $(\tilde{S}_m{}^\varphi)$ defined as follows:

$$\tilde{S}_m{}^\varphi = \{z_k(r_m), w^\mu(z_k(r_m)), w^\gamma(z_k(r_m)) \mid z_k \in [0,1]\} \qquad (11)$$

where $\varphi = 1, \ldots, \Phi$, and $\Phi$ is the number of different generated IFSs.

*Step 5: iFCM Construction*

This step defines the structure of the iFCM model, including the determination of its concepts and relations. Unlike conventional FCMs, which require the participation of experts, the introduced framework embeds a learning algorithm for automatic determination of its structure and the weights of its interconnections based on the training dataset.

Considering the graph nodes of the iFCM, the nodes representing the **input concepts** represent the clusters with medoids $r_m$ of each examined class $f = 1, \ldots, F$ (step 2, step 3), and the **output concepts** represent the different classes $f = 1, \ldots, F$ included in the training images of the dataset used. Figure 4 illustrates an iFCM model defined using the proposed framework, for the example of Fig. 3. The model contains six input concepts, *i.e.*, $C_1 - C_6$, corresponding to the six clusters with medoids $r_m, m = 1, \ldots 6$ of that example, and three output concepts because the classification problem involves three classes (Fig. 3).

The influence between two related concepts $C_j$, $C_i, i \neq j$, $i, j = 1, \ldots, N$, where $N = \mathcal{M} + F$ is the total number of concepts, is expressed by an IFS of the form $\{\langle w^\mu, w^\gamma \rangle\}_{ji}$. This IFS is created following the process described in step 4. Two types of relations can be distinguished: the relations between a) input-output concepts, and b) input concepts. Figure 5 illustrates subgraphs of the FCM shown in Fig. 4, to demonstrate the calculation of the weights of the interconnections between the concepts, which is detailed in the following paragraphs.

**Relations between input and output concepts**. To define and linguistically characterize the influence between such concepts, IFSs $\tilde{S}_{ji} = \{\langle w^\mu, w^\gamma \rangle\}_{ji}$, are constructed as follows:

$$\tilde{S}_{ji} = \bigcup_{\varphi=1}^{\Phi} \tilde{S}_m^\varphi \qquad (12)$$

where $\tilde{S}_m^\varphi \subseteq \tilde{B}_m^e \times \tilde{Q}_m^e, m = 1, \ldots, M$. The symbol "∪" represents the *intuitionistic* fuzzy union operation performed between $C_j$ and $C_i$, $i, j = 1, \ldots, N, i \neq j$. Based on (Atanassov, 1994), the *intuitionistic* fuzzy union operation is performed by aggregating the fuzzy sets $\tilde{B}_m^e$ (which are used for the definition of the IFS memberships) using the fuzzy union operation, and the fuzzy sets $\tilde{Q}_m^e$ (which are used for the definition of the IFS non-memberships) with the fuzzy intersection. The rationale behind using the *intuitionistic* fuzzy union operation comes from the process that occurs in the construction of conventional FCMs, where all expert opinions are taken into account to model the relations that exist between various factors in a given problem.

In Fig.5, examples of such relations are $\{\langle w^\mu, w^\gamma\rangle\}_{17}, \{\langle w^\mu, w^\gamma\rangle\}_{18}, \{\langle w^\mu, w^\gamma\rangle\}_{19}$. Regarding $\{\langle w^\mu, w^\gamma\rangle\}_{17}$, where $C_1$ corresponds to the cluster with medoid $r_1$ of class 1 ($f = 1$), and $C_7$ represents the output concept class 1 ($f = 1$) (Fig. 4). To determine the relation between $C_1$ and $C_7$, a set of IFSs ($\tilde{S}_1^\varphi, \varphi = 1, \ldots, \Phi$) where $\tilde{S}_1^\varphi \subseteq \tilde{B}_1^e \times \tilde{Q}_1^e$. The IFS corresponding to the relation between $C_1$ and $C_7$ is $\tilde{S}_{17} = \{\langle w^\mu, w^\gamma\rangle\}_{17}$ and is used to characterize linguistically the similarities $z_1(r_1)$ to $z_K(r_1)$ (framed in a blue box in Fig. 5).

**Relations between input concepts**. To linguistically characterize the influences between the input concepts, respective IFSs are defined as follows:

$$\tilde{S}_{ji} = \left(\bigcup_{\varphi=1}^\Phi \tilde{S}_{m_i}^\varphi\right) \cap \left(\bigcup_{\varphi=1}^\Phi \tilde{S}_{m_j}^\varphi\right) \qquad (13)$$

where $\tilde{S}_{m_i}^\varphi \subseteq \tilde{B}_{m_i}^e \times \tilde{Q}_{m_i}^e$ and $\tilde{S}_{m_j}^\varphi \subseteq \tilde{B}_{m_j}^e \times \tilde{Q}_{m_j}^e$, $m = 1, \ldots, M, i, j = 1, \ldots, N, i \neq j$. The symbol "∩" represents the *intuitionistic* fuzzy intersection operation performed between $C_j$ and $C_i$. The fuzzy sets $\tilde{B}_{m_i}^e, \tilde{B}_{m_j}^e$ are aggregated, using the fuzzy union operation, while for the fuzzy sets $\tilde{Q}_{m_i}^e$ and $\tilde{Q}_{m_j}^e$ the fuzzy intersection is used, resulting in $\bigcup_{\varphi=1}^\Phi \tilde{S}_{m_i}^\varphi$ and $\bigcup_{\varphi=1}^\Phi \tilde{S}_{m_j}^\varphi$. Based on (Atanassov, 1994), the *intuitionistic* fuzzy intersection operation is performed by aggregating the resulting IFSs, as shown in (13). The rationale behind using the fuzzy intersection is that it represents the consensus among expert opinions that exist when determining the relations in conventional FCMs.

In Fig. 5 indicative examples of such relations are $\{\langle w^\mu, w^\gamma\rangle\}_{21}, \{\langle w^\mu, w^\gamma\rangle\}_{23}$. Regarding $\{\langle w^\mu, w^\gamma\rangle\}_{23}$, $C_2$ corresponds to the cluster with medoid $r_2$ of class 1 ($f = 1$), and $C_3$ corresponds to the cluster with medoid $r_3$ of class 2 ($f = 2$). To determine the relation between $C_2$ and $C_3$, sets of IFSs $\{\tilde{S}_2^{\varphi_2} | \varphi_2 = 1, \ldots, \Phi_2\}$ and $\{\tilde{S}_3^{\varphi_3} | \varphi_3 = 1, \ldots, \Phi_3\}$, where $\tilde{S}_2^{\varphi_2} \subseteq \tilde{B}_2^e \times \tilde{Q}_2^e$ and $\tilde{S}_3^{\varphi_3} \subseteq \tilde{B}_3^e \times \tilde{Q}_3^e$. Finally, the IFS corresponding to the relation between $C_2$ and $C_3$ is given by $\tilde{S}_{23} = \{\langle w^\mu, w^\gamma\rangle\}_{23}$. In addition, $\tilde{S}_{23}$ is used to linguistically characterize the similarities $z_1(r_2)$ to $z_K(r_2)$ and $z_1(r_3)$ to $z_K(r_3)$ (framed in a blue box in Fig.5).

In all the above-mentioned cases, the interconnection weights between concepts $C_j$ and $C_i$ for every $j \neq i$ are calculated using the intuitionistic center of area of each $\tilde{S}_{ji}$ (Angelov, 1995) for the number $\xi = card(\{I_k | I_k \text{ belongs to class } f\})$ of images $I_k, k = 1, \ldots, K$ belonging to a class $f$

$$z_{ji} = \frac{\sum_{k=1}^\xi \left(\left(w_{ji}^\mu(z_k(r_m)) - w_{ji}^\gamma(z_k(r_m))\right) \cdot z_k(r_m)\right)}{\sum_{k=1}^\xi \left(\left(w_{ji}^\mu(z_k(r_m)) - w_{ji}^\gamma(z_k(r_m))\right)\right)} \qquad (14)$$

where $w_{ji}^\mu(z_k(r_m)) > w_{ji}^\gamma(z_k(r_m))$. Based on the above, the relations of $C_j$ to $C_i$ for every $j \neq i$ are described with weight pairs $\langle w_{ji}^\mu(z_{ji}), w_{ji}^\gamma(z_{ji})\rangle$.

### 4.2 Test Phase

Given a test input image $I$, feature vectors $\boldsymbol{D} = \{d^1, \ldots, d^{P'}\}$ are extracted using a CNN, as described in step 1 of the training phase. The similarity $z(r_m)$ between the calculated medoids $r_m$, $m = 1, ..., M$ and $D = \{d^1, ..., d^{P'}\}$ is calculated, and they are linguistically characterized using IFSs (steps 3,4). In this way, the state vector of the form:

$$A = (\langle w^\mu(z(r_1)), w^\gamma z(r_1)\rangle, ..., \langle w^\mu(z(r_M)), w^\gamma z(r_M)\rangle) \quad (15)$$

of the image is calculated and used as input in the trained iFCM model, aiming to perform interpretable classification. The reasoning process is then performed by iteratively calculating the pairs $\langle v^\mu, v^\gamma \rangle_i$ of each $C_i, i = 1, ..., N$ using (6)-(8).

## 5. Experiments and Results
### 5.1 Datasets and Settings

To evaluate the performance of I²FCM, the following publicly available benchmark datasets were used:

a) **Caltech 101** (Fei-Fei et al., 2004) contains 9,144 images of 102 different classes. Each class includes 33 to 800 images, and the size of the images on average is 300 × 200 pixels. For the experiments, 30 randomly selected images from each class of the dataset were used as the training set while the rest images formed the test set.

b) **Caltech 256** (Griffin et al., 2007) consists of 30,607 images, illustrating 256 different object classes with at least 80 images in each class. The images of the dataset have an average resolution of 300 × 250 pixels. For the evaluation, 60 randomly selected images from each class composed the training set, and the rest of them are used for testing.

c) **15-Scenes** (Lazebnik et al., 2006) includes 200 to 400 images of 15 classes with an average resolution of 300 × 250 pixels. A set of 100 images was used for training and the rest of them for testing.

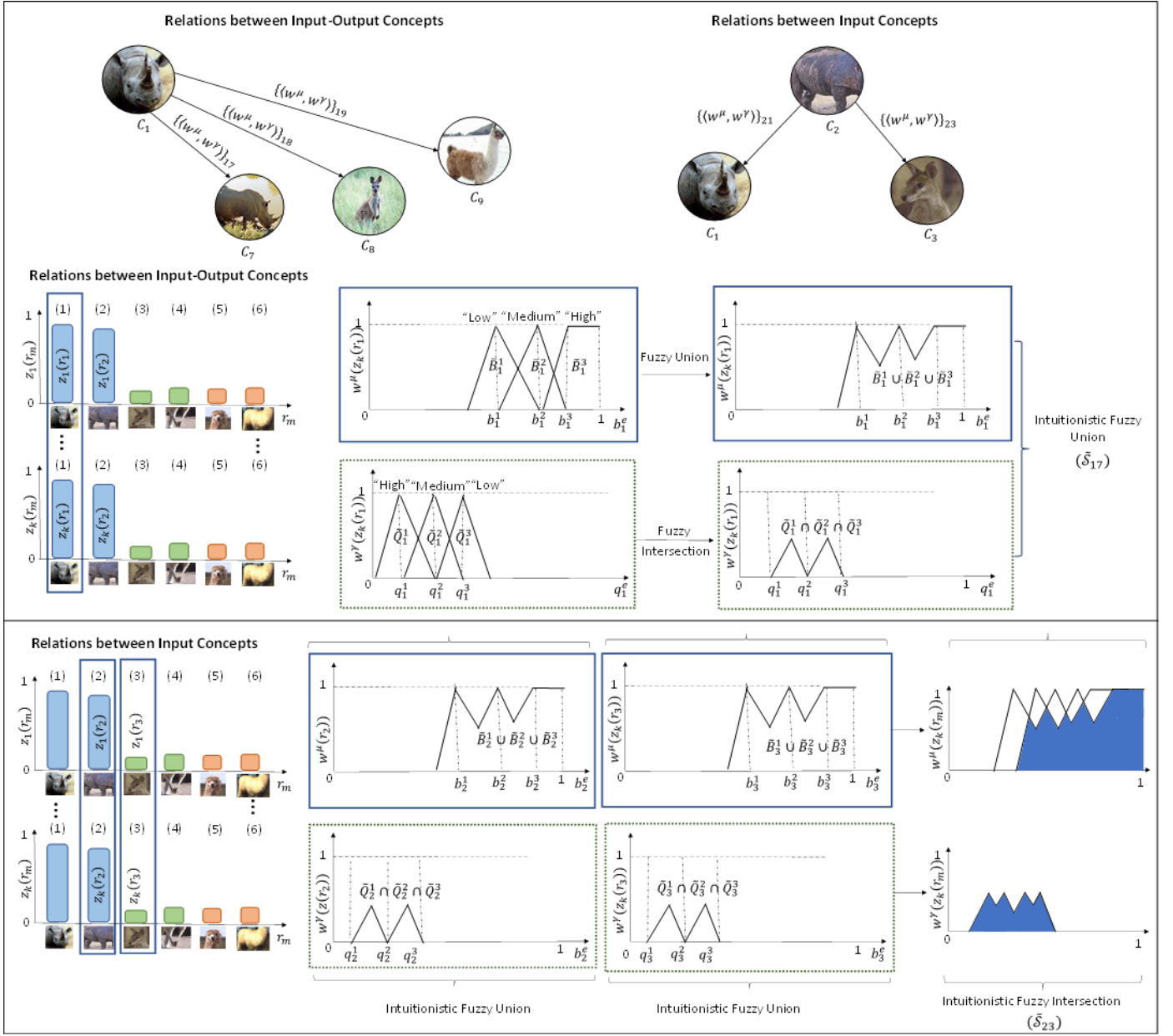

**Fig. 5** Weight definition based on I²FCM (step 5).

The feature extraction process (step 1) was performed using VGG-16 model pretrained on ImageNet (Deng et al., 2009). The clustering process was implemented using the *k*-means algorithm (Drake and Hamerly, 2012) with Euclidean distance, as a baseline clustering algorithm commonly used in related methods (Vasilakakis and Iakovidis, 2023). The range of the number of clusters in *k*-means tested using grid search was [5, 50], and the range of number of fuzzy sets tested using grid search was [5, 15].

The fuzzy sets were implemented using Gaussian membership functions. In addition, the number of iFCM input concepts ranged from 2 to 15, and the number of output concepts were defined according to the number of classes included in the dataset used. The hesitancy values of the iFCM concepts were initially set to 0. Furthermore, the hyperbolic tangent (tanh) was used as a transfer function in $\{\langle \mathcal{F}_{\mu_s}, \mathcal{F}_{\gamma_s}\rangle\}$ of (6)-(7).

**Table 2** Classification Accuracy on Caltech-101 and Caltech-256.

| Models | Caltech-101 | Models | Caltech-256 |
|---|---|---|---|
| VGG-16 | 0.903 | VGG-16 | 0.732 |
| ResNet-50 | 0.903 | VGG-19 | 0.706 |
| k-NN | 0.849 | ResNet-101 | 0.751 |
| DRB | 0.845 | GoogLeNet | 0.724 |
| xDNN | 0.943 | xDNN | 0.754 |
| H-DRB | 0.895 | H-DRB | 0.731 |
| DMR | 0.943 | DMR | 0.775 |
| xFCM | 0.910 | xFCM | 0.736 |
| **I$^2$FCM** | 0.920 | **I$^2$FCM** | 0.752 |

**Table 3** Classification Accuracy on 15-Scenes.

| Models | 15-Scenes |
|---|---|
| VGG-16 | 0.906 |
| k-NN | 0.812 |
| DAG-CNN | 0.929 |
| xFCM | 0.912 |
| **I$^2$FCM** | 0.934 |

### 5.2 Classification Performance

The performance of I$^2$FCM, was compared with both state-of-the-art and conventional machine learning models in terms of classification accuracy. Comparisons were performed with the following models:

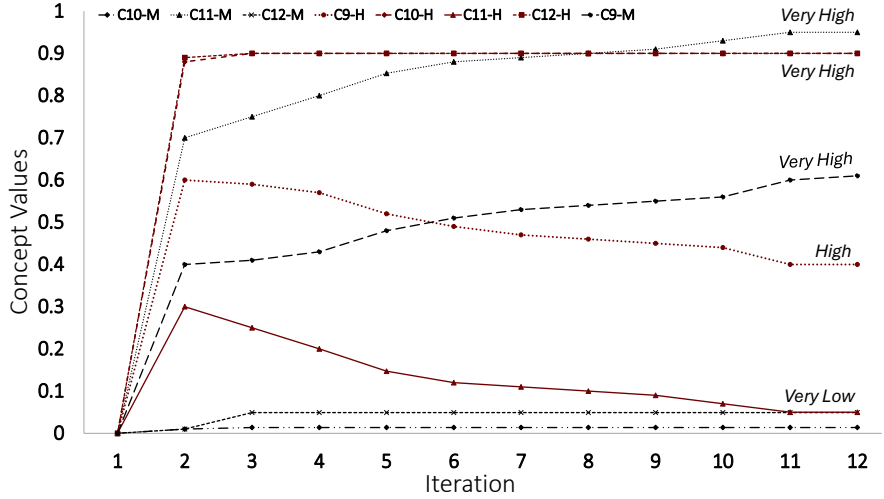

**Fig. 6.** Convergence of iFCM during reasoning Membership (-M) and Hesitancy (-H).

a) VGG-16 (Simonyan and Zisserman, 2014), b) VGG-19 (Simonyan and Zisserman, 2014), c) GoogleLeNet (Szegedy et al., 2015), d) ResNet-101 (He et al., 2016), e) ResNet-50 (He et al., 2016), f) Directed Acyclic Graph (DAG)-CNN (Yang and Ramanan, 2015), g) *k*-Nearest Neighbors (k-NN) algorithm (Koutroumbas and Theodoridis, 2008) as well as relevant state-of-the-art interpretable models: a) DMR (Angelov and Soares, 2020b), b) xDNN (Angelov and Soares, 2020a), c) H-DRB (Gu and Angelov, 2020), d) DRB (Angelov and Gu, 2018), and e) xFCM (Sovatzidi et al., 2022a).

The results are summarized in Tables 2-3, where the interpretable classification methodologies are underlined. It can be noticed that I²FCM outperforms most of the compared models in terms of the classification accuracy on all datasets. It is worth noting that I²FCM outperforms all non-interpretable models. More importantly, comparing its classification performance with that of the original VGG-16 network, it becomes evident that its use enhances the classification of the same feature vectors (since VGG-16 is used for feature extraction in step 1 (section 3). Some of the compared interpretable methods in Table 2 perform somewhat better than I²FCM; however, as noted in Table 1, the latter has the advantage of providing interpretations that include semantically relevant details about image contents. To further highlight this aspect, the following subsection provides an analysis of the interpretability of I²FCM with a specific example.

### 5.3 Interpretability Analysis

Let us consider a test input image, which depicts a flamingo from the Caltech-101 dataset. According to step 1 of the proposed framework feature vectors $\boldsymbol{D}$ are extracted from $K = 9$ selected superpixels. The iFCM model constructed in step 5 of the training process consists of a total of $\mathcal{M} = 8$ input concepts are used, which are the following: $C_1$: "Rhino-head", $C_2$: "Rhino-body", $C_3$: "Kangaroo-head", $C_4$: "Kangaroo-body", $C_5$: "Flamingo-head", $C_6$: "Flamingo-body", $C_7$: "Llama-head", $C_8$: "Llama-body". The output concepts represent the corresponding $F = 4$ classes that are: $C_9$: "Rhino", $C_{10}$: "Kangaroo", $C_{11}$: "Flamingo", and $C_{12}$: "Llama". The iFCM model has relations between all input concepts, as well as relations between input and output concepts. To simplify its representation in Fig. 7(a) only a subset of the relations is depicted. The fuzzy sets corresponding to the linguistic values of the influences between its concepts are illustrated in Fig. 7 (b).

The initial state vector of the input image is calculated (after steps 2 and 3) and is utilized as input to the iFCM model. Then, the iFCM iteratively calculates its concept values, until it reaches a steady state, in $t = 12$ iterations, using (6)–(8). The convergence of the output concepts of the iFCM model, in terms of its membership and hesitancy values, is illustrated in Fig. 6. Also, the results of the image classification after $t = 12$ iterations, are presented in Fig. 7(d). In this figure, green bars represent membership, blue bars represent non-membership, and yellow bars represent the respective hesitancy values. It can be observed that in all cases, the non-membership values reach zero at the steady state. This is due to the reasoning performed using (7) (Papageorgiou and Iakovidis, 2012). Thus, the initial non-membership values do not practically affect the final decision. The hesitancy values are decreased for the iFCM nodes containing parts characterizing the class to which the image belongs to. In this example nodes $C_5, C_6$ contain parts characterizing the class and $C_{11}$: "Flamingo" (Fig. 7(d)). In case an image belongs to a certain class with high similarity with the corresponding concepts, the membership function tends to 1, while the non-membership and hesitancy values tend to 0. The low degree of hesitancy confirms the validity of the decision.

To summarize, the explanation about the classification outcome for the test input image, after $t = 12$ iterations, is the following:

- The input image is classified as "Flamingo" because it has (a) Very High similarity with Very Low hesitancy with $C_5$ ="Flamingo-head"; (b) Very High similarity with Very Low hesitancy with $C_6 =$ "Flamingo-body".

- The input image cannot be classified as "Rhino" because it has (a) Low similarity with a Very High hesitancy with $C_1$: "Rhino-head", and (b) High similarity with Very High hesitancy with $C_2$: "Rhino-body".

The high degree of similarity, *i.e.*, membership value, in the latter case can be explained by the fact that the extracted features vectors $d^{10}, d^{12}, d^{15}$ corresponding to the background of the image (Fig. 7(a)), have a similar visual content with $C_2$ that represents the "Rhino-body". More specifically, as illustrated in the figure, in the test image there are regions of dark blue in the background, which are similar to the body of rhino.

Regarding the relations of iFCM for the image classification problem under consideration, the following conclusions can be derived: for the membership values there is a *positive* causality (whereas for the non-membership values there is a *negative* causality) between:

(a) input and output concepts that are related to the same class. For example, there is a positive relation between $C_1$ and $C_2$, representing "Rhino-head" and "Rhino-body", respectively. This means that as the similarity of the extracted features from the test input image with the representative features depicting "Rhino-head" increases, there is also an increase to the similarity with "Rhino-body".
(b) input and output concepts related to the same class. For example, Fig. 7 (a) shows that there is a positive influence of the input concepts $C_1$ and $C_2$, with the output concept $C_9$ belonging to the class "Rhino".

In addition, there is a negative causality in the membership values (whereas there is a positive causality in the non-membership values), among:

- the input concepts related to different classes.
- the input and output concepts representing different classes.

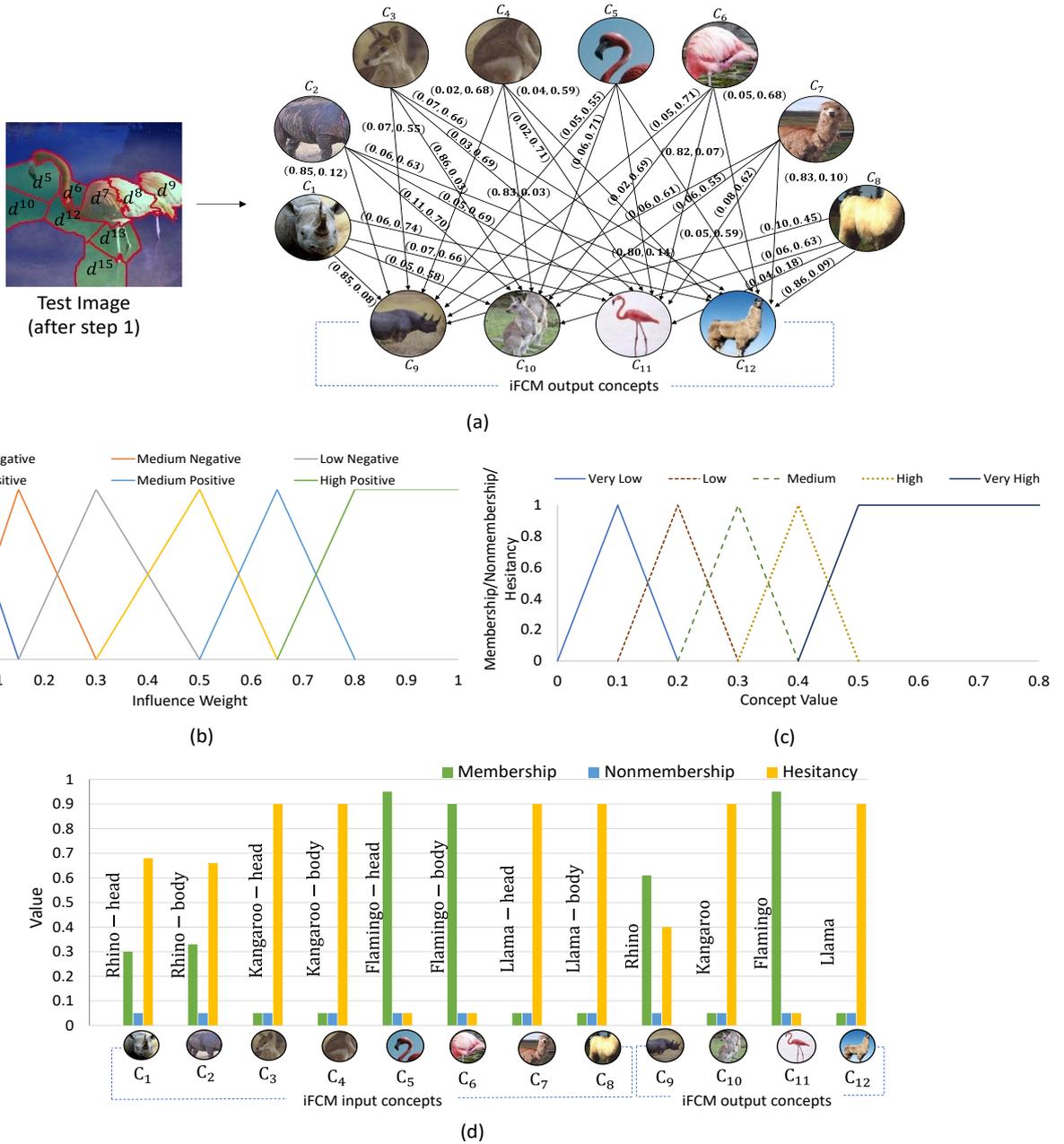

**Fig. 7** (a) Trained iFCM model, (b) linguistic values for influence weights, (c) linguistic values for membership, non-membership and hesitancy values for concepts $C_1 - C_{12}$, and (d) membership, non-membership, hesitancy values for concepts $C_1 - C_{12}$ at $t = 12$.

## 6. Discussion and Conclusions

In this paper, a novel framework named I2FCM is introduced. The proposed framework constructs iFCM models for interpretable image classification, and to the best of our knowledge it is the first iFCM-based approach applied in the context of image analysis. The main characteristic of iFCMs is their natural mechanism to assess the quality of their output through the estimation of hesitancy.

The application of I2FCM on well-known benchmark datasets showed that it is competitive to state-of-the-art in terms of classification performance, while providing meaningful explanations. Compared to the rest interpretable models, *i.e.,* DMR (Angelov and Soares, 2020b), xDNN (Angelov and Soares, 2020a), HDRB (Gu and Angelov, 2020), DRB (Angelov and Gu, 2018), enables the user to monitor the decision-making process while at the same time explaining the result obtained at each iteration, in an understandable way, using the causality existing among the graph nodes. In this way, it strengthens the confidence of the user, and makes the framework useful and suitable to be implemented in highly sensitive areas, including healthcare (Angelov et al., 2021). Furthermore, compared to xFCM (Sovatzidi et al., 2022a), the proposed I$^2$FCM framework enables the image interpretation based on local visual content captured by selective superpixel-based spatial sampling of image regions.

In addition, the proposed framework defines iFCMs that are able to distinguish representative semantics among various image classes, and their cause-and-effect relations. The evaluation of the iFCM using publicly available benchmark datasets indicated its superiority over other state-of-the-art methods, including interpretable ones.

Advantages of I$^2$FCM over current relevant classification frameworks, include the following:

- Intuitionistic fuzzy logic improves the knowledge elicitation and, consequently, the decision-making process of FCMs.
- Includes an algorithm for automatic determination of the graph structure from the data used, thus limiting human intervention and participation of experts.
- It provides more accurate outcomes than most other interpretable state-of-the-art approaches compared, and this can be attributed to the improved modeling of the uncertainty through IFSs.
- It can encapsulate CNN classifiers, thus making them interpretable.
- It is simple to implement.

Future directions include further investigation of the proposed framework, including its performance on various application domains.